\documentclass[manuscript]{acmart}

\usepackage[utf8]{inputenc} 
\usepackage[T1]{fontenc}    
\usepackage{hyperref}       
\usepackage{url}            
\usepackage{booktabs}       
\usepackage{amsfonts}       
\usepackage{amsmath}       
\usepackage{nicefrac}       
\usepackage{microtype}      
\usepackage{xcolor}         
\usepackage{hyperref}
\usepackage{graphicx}
\usepackage{subcaption}
\usepackage{xspace}
\newcommand{\carbondioxide}{CO\textsubscript{2}\xspace}
\newcommand{\carboneq}{CO\textsubscript{2}eq\xspace}
\newcommand{\carbonintensity}{gCO\textsubscript{2}eq/kWh\xspace}

\AtBeginDocument{%
  \providecommand\BibTeX{{%
    \normalfont B\kern-0.5em{\scshape i\kern-0.25em b}\kern-0.8em\TeX}}}
\setcopyright{none}
    
\begin{document}
\title{Counting Carbon: A Survey of Factors Influencing the Emissions of Machine Learning}

\author{Alexandra Sasha Luccioni}
\affiliation{
\institution{Hugging Face}
\country{Montreal, Canada}}
\email{sasha.luccioni@huggingface.co}
\author{Alex Hernandez-Garcia}
\affiliation{
\institution{Mila, Université de Montréal}
\country{Montreal, Canada}}
\email{alex.hernandez-garcia@mila.quebec}

\begin{abstract}
Machine learning (ML) requires using energy to carry out computations during the model training process. The generation of this energy comes with an environmental cost in terms of greenhouse gas emissions, depending on quantity used and the energy source.  Existing research on the environmental impacts of ML has been limited to analyses covering a small number of models and does not adequately represent the diversity of ML models and tasks. In the current study, we present a survey of the carbon emissions of 95 ML models across time and different tasks in natural language processing and computer vision. We analyze them in terms of the energy sources used, the amount of \carbondioxide~emissions produced, how these emissions evolve across time and how they relate to model performance. We conclude with a discussion regarding the carbon footprint of our field and propose the creation of a centralized repository for reporting and tracking these emissions.
\end{abstract}

\maketitle
\section{Introduction}~\label{sec:introduction}

In recent years, machine learning (ML) models have achieved high performance in a multitude of tasks such as image classification, machine translation, and object detection. However, this progress also comes with a cost in terms of energy, since developing and deploying ML models requires access to computational resources such as Graphical Processing Units (GPUs) and therefore energy to power them. In turn, producing this energy comes with a cost to the environment, given that energy generation often entails the emission of greenhouse gases (GHG) such as carbon dioxide (\carbondioxide)~\cite{rodhe1990comparison}. On a global scale, electricity generation represents over a quarter of the global GHG emissions, adding up to 33.1 gigatonnes of \carbondioxide in 2019~\cite{energy2019co2}. 
Recent estimates put the contribution of the information and communications technology (ICT) sector -- which includes the data centers, devices and networks used for training and deploying ML models -- at 2--6~\% of global GHG emissions, although the exact number is still debated~\cite{itu2020greenhouse,malmodin2018energy, ict2020}. In fact, there is limited information about the overall energy consumption and carbon footprint of our field, how it is evolving, and how it correlates with performance on different tasks.

The goal of the current paper is to analyze the main factors influencing the carbon emissions of our field, to study the evolution across time, and to contribute towards a better understanding of the carbon emissions generated by ML models trained on different tasks and as a function of their performance. As such, our research aims to answer the following research questions:
\begin{enumerate}
    \item What are the main sources of energy used for training ML models?
    \item What is the order of magnitude of \carbondioxide emissions produced by training ML models?
    \item How do the \carbondioxide~emissions produced by training ML models evolve over time?
    \item Does more energy and \carbondioxide lead to better model performance?
\end{enumerate}

We start our article with a survey of related work in Section~\ref{related}, followed by a presentation of our methodology in Section~\ref{methods}. In Section~\ref{sec:results} we present our analysis, and we conclude with our proposals for future work, including a centralized hub for reporting the carbon footprint of machine learning..

\section{Related work}~\label{related}
Measuring the environmental impact of ML models is a relatively new undertaking, but one that has been gathering momentum in recent years. In the current section, we present several directions pursued in this domain, from empirical studies of specific models to the development of efficient algorithms and hardware. 

\paragraph{Empirical studies on carbon emissions} A large proportion of research has focused on estimating the carbon emissions of specific model architectures and/or comparing the carbon emissions of two or more models and approaches. The first paper to do so was written by Strubell et al., which estimated that the emissions of training and fine-tuning a large Transformer model with Neural Architecture Search (NAS) produced 284,019 kg (626,155 lbs) of \carbondioxide, similar to the lifetime emissions of five US cars.~\cite{strubell2019energy}. This perspective has since been explored further via analyses of the carbon footprint of different neural network architectures~\cite{luccioni2022estimating, patterson2021carbon,patterson2022carbon} and the relative efficiency of different methods~\cite{yusuf2021curb,naidu2021towards}. These empirical studies are very recent (post-2019),  remain relatively sparse and biased towards certain research areas (i.e. Natural Language Processing), and there are many aspects of the emissions of model training that remain unexplored. In sum, there is a need for a more broad and multi-faceted analysis in order to better understand the scale and variation of carbon emissions in our community.

\paragraph{Tools and approaches for measuring carbon emissions} Developing standardized approaches for estimating the carbon emissions of model training has also been the focus of much work~\cite{lottick2019energy,schmidt2021codecarbon,henderson2020towards,lannelongue2021green,lacoste2019quantifying, trebaol2020cumulator, anthony2020carbontracker}. As a result, there are several tools that exist for this purpose, such as \href{https://github.com/mlco2/codecarbon}{Code Carbon} and the \href{https://github.com/Breakend/experiment-impact-tracker}{Experiment Impact Tracker}, which can be used during the model training process, or the \href{https://mlco2.github.io/impact/}{ML CO2 Calculator}, which can be used after training, all of which provide an estimate of the amount of carbon emitted. However, a recent study on different carbon estimation tools concluded that the estimates produced by different tools vary significantly and consistently under-report emissions~\cite{bannour2021evaluating}. To date, there is no single, accepted approach for estimating the carbon emissions of the field, making standardized reporting and comparisons difficult~\cite{luccioni2022estimating}.

\paragraph{Broader impacts of ML models} Several papers have been written in recent years regarding the broader societal impacts of ML models, which includes their environmental footprint. This spans research on how the size and computational demands of ML models in general~\cite{thompson2020computational} and large language models in particular~\cite{bender2021dangers,bommasani2021opportunities}  have grown in recent years. Many strategies and directions forward have been proposed, ranging from advocating for more environmentally-conscious practice of AI~\cite{schwartz2020green} to adopting a sustainability mindset for the community~\cite{wu2021sustainable}. However, while the documentation of aspects such as bias and safety has begun to be described in reports and articles accompanying certain recent ML models (e.g.~\cite{brown2020language,chinchilla2022}), environmental impacts have yet to be consistently tracked and reported. Notable exceptions include recent language models such as OPT~\cite{opt2022}, T0~\cite{t0-2021} and BLOOM~\cite{luccioni2022estimating}.

\paragraph{Efficient algorithms and hardware} A related and complementary direction of research is the development of more efficient model architectures and approaches. For instance, approaches such as Eyeriss~\cite{chen2019eyeriss} and DistilBERT~\cite{sanh2019distilbert} have made significant progress in terms of computing efficiency, enabling faster training and inference, which results in less energy usage and, indirectly, less carbon emissions, during model training. This research is gathering attention within the community, with workshops such as \href{https://sites.google.com/view/sustainlp2022/home}{SustaiNLP} and \href{https://www.emc2-ai.org/neurips-19}{EMC2} growing in scope and popularity, although efficiency has yet to be a central consideration when it comes to evaluating and comparing models. However, energy-efficient benchmarks such as HULK~\cite{zhou2020hulk} have also been proposed, which take computational requirements and environmental impacts into account during model evaluation, allowing a comparison of models based on multiple criteria. 

\paragraph{Other aspects of the carbon impact of ML} Finally, efforts have been made to quantify other factors that have an influence on the overall carbon footprint of the field of ML, including in-person versus virtual conference attendance~\cite{skiles2021conference}, the manufacturing of computing hardware~\cite{gupta2021chasing}, life cycle analysis of the entire ML development and deployment cycle~\cite{ligozat2021unraveling}, as well as some initial studies regarding the carbon footprint of model deployment in production settings~\cite{luccioni2022estimating}. The relative contribution of each of these factors is still unclear, which suggests that further research is needed in order to further disentangle these factors.
    
\section{Methodology} \label{methods}

As stated in Section~\ref{sec:introduction}, the goal of this paper is descriptive -- to observe the evolution of the carbon emissions of our field of ML across time and to analyze the different aspects of the carbon emissions produced by training ML models. In this section, we present the different aspects and details of our methodology.

\subsection{Data collection} 

In order to gather data from a diverse set of ML models from a variety of domains and tasks, we leveraged the dataset collected by Thompson et al.~\cite{thompson2020computational} in the scope of a recent study on the computational requirements of ML. From this dataset, we equally sampled 500 papers published from 2012 to 2021 spanning 5 tasks: Image Classification, Object Detection, Machine Translation, Question Answering and Named Entity Recognition. We then contacted the first author of each of the papers and asked them to provide missing training details regarding their model (See Supplementary Materials~\ref{appendix:emails} for the email text). We were able to collect information for a total of 95 models from 77 papers (since some of the papers trained more than one model), which represents an author response rate of 15.4~\%.

\begin{table}[h!]
\caption{Summary of the models analyzed in our study}
\begin{center}
\hspace{-1cm}
\begin{tabular}{lccc}
\toprule
\textbf{Task} & \textbf{Dataset} & \textbf{Number of Models} & \textbf{Publication dates} \\
\midrule
Image Classification  &    ImageNet~\cite{deng2009imagenet}      & 35    & 2012-2021     \\
Machine Translation    &   WMT2014~\cite{bojar2014findings}      & 30 & 2016-2021  \\
Named Entity Recognition &  CoNLL 2003~\cite{sang2003introduction}  & 11    & 2015-2021  \\
Question Answering & SQuAD 1.1~\cite{rajpurkar2016squad} & 10 & 2016-2021 \\
Object Detection & MS COCO~\cite{lin2014microsoft} & 9 & 2019-2021 \\
\bottomrule
\end{tabular}
\label{table:papers}
\end{center}
\end{table}

The models in our sample cover a diversity of tasks spanning nine years of research in the field and a variety of conferences and journals. They all represent novel architectures at the time of publication, achieving high performance in their respective tasks: on average, the models are within 8~\% of SOTA performance according to \href{https://paperswithcode.com/}{Papers With Code} leaderboards 
  at the time of their publication
. This sample represents the largest amount of information regarding the carbon footprint of ML model training to date, and provides us with opportunities to analyze it from a variety of angles, which we present in Section~\ref{sec:results}. In the remaining of this section, we describe our method for estimating carbon emissions.

\subsection{Estimating carbon emissions}

The unit of measurement typically used for quantifying and comparing carbon emissions is \emph{\carbondioxide~equivalents}. This unit allows us to compare different sources of greenhouse (GHG) emissions using a common denominator, that of grams of \carbondioxide~ emitted per kilowatt hour of electricity generated (\carbonintensity)~\footnote{For instance, methane is 28 times more potent than \carbondioxide~based on its 100-year global warming potential, so energy generation emitting 1 gram of methane per kWh will emit 28 grams of \carboneq~per kWh.}. 

The amount of \carboneq~($C$) emitted during model training can be decomposed into three relevant factors: the power consumption of the hardware used ($P$), the training time ($T$) and the carbon intensity of the energy grid ($I$); or equivalently, the energy consumed ($E$) and the carbon intensity:
\begin{equation}
 C = P \times T \times I = E \times I.
\end{equation}
For instance, a model trained on a single GPU consuming 300~W for 100 hours on a grid that emits 500~\carbonintensity~will emit $0.3~\text{kW} \times 100~\text{h} \times 500~\text{g/kWh} = 15000~\text{g} = 15~\text{kg}$ of \carboneq. The same model trained on a less carbon-intensive energy grid, emitting only 100~\carbonintensity, will only emit $0.3 \times 100 \times 100 = 3000~\text{g} = 3~\text{kg}$ of \carboneq, i.e. five times less overall. 
In our email to authors, we asked them to provide the details we needed to carry out this calculation, i.e the location of the computer or server where their model was trained (either cloud or local), the hardware used, and the total model training time. We describe how we estimate each of the relevant factors in the paragraphs below:

\paragraph{Carbon Intensity} Based on the training location provided by authors, we were able to estimate the carbon intensity of the energy grid that was utilized, based on publicly-available sources such as the \href{https://www.iea.org/}{International Energy Agency} and the \href{https://www.eia.gov/}{Energy Information Administration}. The granularity of information available ranges widely depending on the location -- whereas in countries such as the United States, it is available at a sub-state (sometimes even at a sub-zip code) level, in others such as China, only country-level information is available. The carbon intensity figures that we use are yearly averages for the year the model was trained, given that these can evolve over time. In cases when the authors indicated that they used a computing infrastructure internal to a company, we consulted company reports and publications (e.g.~\cite{patterson2021carbon, facebook2020}) to obtain more precise information regarding the carbon intensity, including the usage of local renewable energy sources. In cases when models were trained on commercial cloud computing platforms such as Google Cloud or Amazon Web Services (AWS), we used the information provided by the companies themselves to estimate emission factors~\cite{google,aws}.

\paragraph{Hardware power} In order to calculate the power consumption of the hardware used for model training, we refer to its Thermal Design Power, or TDP, which indicates the energy it needs under the maximum theoretical load. That is, the higher the TDP, the more power is consumed. While in practice GPUs are not always fully utilized during all parts of the training process, gathering more precise information regarding real-time power consumption is only possible by using a tool like Code Carbon during the training process~\cite{schmidt2021codecarbon}.  
Nonetheless, the TDP-based approach is often used in practice when estimating the carbon emissions of AI model training~\cite{patterson2021carbon} and it remains a fair approximation of the actual energy consumption of many hardware models. We provide more information about TDP and the hardware used for training the models in our sample in Section~\ref{appendix:hardware} of the Appendix. %
\paragraph{Training Time} Training time was computed as the total number of hardware hours, which is different from the "wall time" of ML model training, since most models were trained on multiple units at once. For instance, if training a model used 16 GPUs for 24 hours, this equals a training time of \emph{384 GPU hours}; a model using 8 GPUs for 48 hours will therefore have an equivalent training time. 

\section{Data analysis} \label{sec:results}

 In the sections below, we present several aspects regarding the carbon footprint of training ML models, examining the main sources of energy used for training (\S~\ref{sec:energymix}), the order of magnitude of \carbondioxide~emissions produced (\S~\ref{sec:ml-emissions}), the evolution of these emissions over time (\S~\ref{sec:carbon-emissions}) and the relationship between carbon emissions and model performance (\S~\ref{sec:task-progress})~\footnote{We have made the data used for our analysis available in a \href{https://github.com/alexhernandezgarcia/co2ml}{GitHub repository}.}. 

\subsection{What are the main sources of energy used for training ML models?} \label{sec:energymix}

The primary energy source used for powering an electricity grid is the single biggest influence on the carbon intensity of that grid, in the face of the large differences between energy sources. For instance, renewable energy sources like hydroelectricity, solar and wind have low carbon intensity (ranging from 11 to 147 \carbonintensity), whereas non-renewable energy sources like coal, natural gas and oil are generally orders of magnitude more carbon-intensive (ranging from 360 to 680 \carbonintensity)~\cite{energy2019co2, schlomer2014annex}. That means that the energy source that powers the hardware to train ML models can result in differences of up to 60 times more \carboneq~in terms of total emissions.

\begin{table}[h!]
\caption{Main Energy Sources for the models analyzed and their carbon intensities~\cite{energy2019co2,eiadata}}
\begin{center}
\begin{tabular}{lccc}
\toprule
\textbf{Main energy source} & \textbf{Number of Models} & \textbf{Low-Carbon?} &
{\begin{tabular}[c]{@{}c@{}}\textbf{Average Carbon Intensity} \\ \textbf{(\carbonintensity)} \end{tabular}}  \\
\midrule
Coal     & 38        & No     & 512.3     \\
Natural Gas   & 23    & No     & 350.5     \\
Hydroelectricity & 19 & Yes    & 100.6       \\
Oil      & 12         & No     & 453.6      \\
Nuclear    & 3      & Yes & 147.2       \\

\bottomrule
\end{tabular}
\label{table:energysource}
\end{center}
\end{table}

In Table~\ref{table:energysource}, we show the principal energy source used by the models from our sample, as well as its average carbon intensity. 
We found that the majority of models (61) from our sample used high-carbon energy sources such as coal and natural gas as their primary energy source. whereas less than a quarter of the models (34) used low-carbon energy sources like hydroelectricity and nuclear energy~\footnote{Although the sustainability of nuclear energy is debated, it is one of the least carbon-intensive sources of electricity that currently exists. More information about nuclear energy and its long-term impacts on the environment can be found in~\cite{apergis2010causal} and ~\cite{suman2018hybrid}.}. While the average carbon intensity used for training the models from our sample (372~\carbonintensity) is lower than the average global carbon intensity (475~\carbonintensity), this still leaves much to improve in terms of carbon emissions of our field by switching to renewable energy sources (we discuss this further in Section~\ref{discussion}).

\begin{figure}[ht!]
  \includegraphics[width=\linewidth]{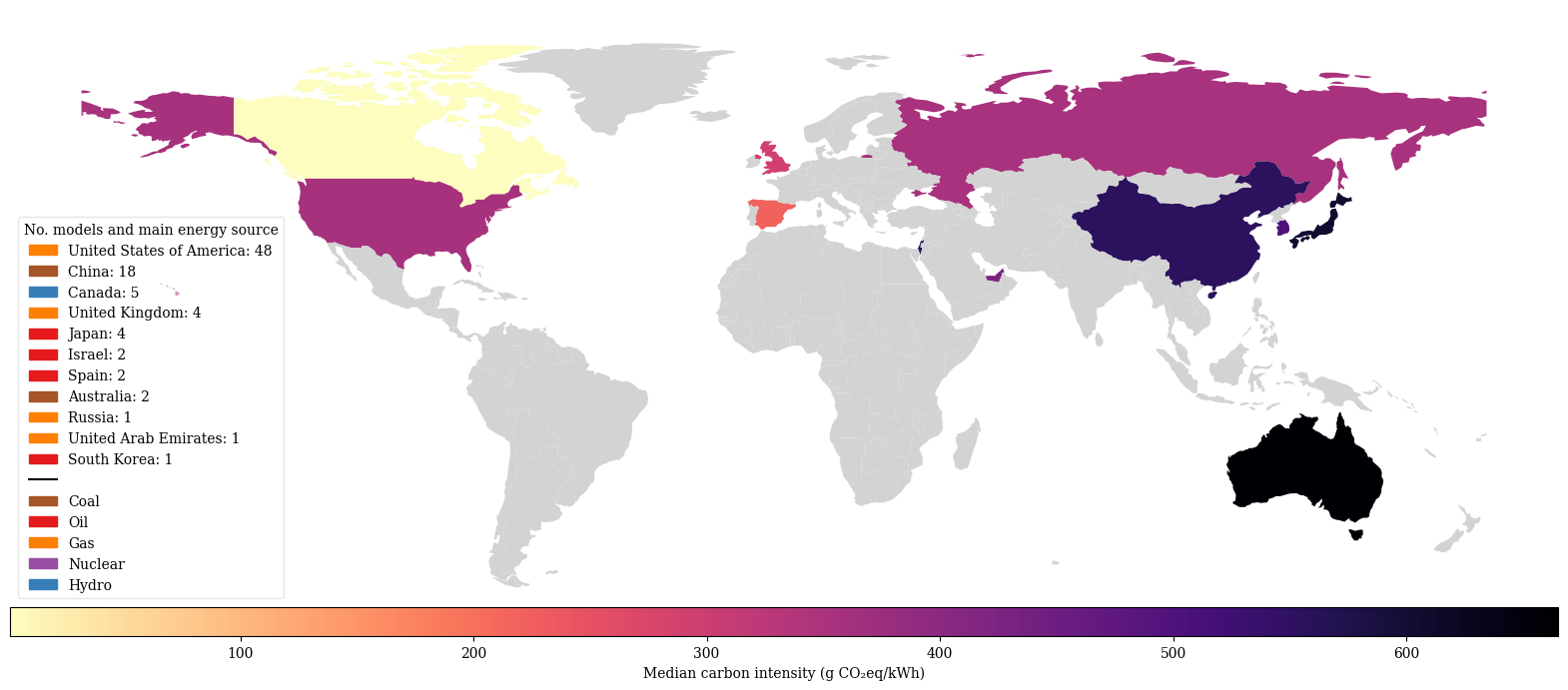}
  \caption{Map with the countries where the models in the data were trained, as reported by the authors. The colors code the median carbon intensity of the energy used by the models trained in each country. The legend indicates the number of models trained in each country, as well as a colored patch marking the main energy source -- see bottom of the legend for the values.}
  \label{fig:map_carbonintensity}
\end{figure}

In Figure~\ref{fig:map_carbonintensity}, we show the model training locations reported by authors on a country-level, with the median carbon intensity of each country indicated below. In terms of the model training locations reported by authors, we found a very imbalanced distribution, with the vast majority of models being trained in a small number of countries -- half of the models in our sample were trained in the United States (48), followed by China (18), with the rest of the models distributed across 9 other countries, with only a few papers in each. Regarding the primary energy sources, based on this country-level analysis of energy grids used for training the models in our sample, we found that most common countries where model training was carried out (e.g. the US and China), are on the high end of the carbon spectrum, with emissions of 350 \carbonintensity~and above. On the other end, the countries with the lowest carbon intensity in our sample are Canada (which ranges between 1.30 and 52.89~\carbonintensity,~depending on the province) and Spain (which has a single national energy grid with a median carbon intensity of 220.26~\carbonintensity), but they only represents a total of 7 models from our sample. This is similar to patterns in emissions worldwide, where a small number of highly industrialized countries produce the majority of the world's greenhouse gases~\cite{friedrich2020interactive}.

Another observation that can be made based on our data is that none of the models from our sample were trained in either Africa nor South America -- in fact, the majority of the models from our sample (76) were trained in countries representing the Global North. This is consistent with previous work examining the `digital divide' in ML and observing the centralization of power in the field, which hinders researchers from underrepresented locations and groups from contributing to the field, given the attribution of computing resources~\cite{birhane2021values, ahmed2020democratization, ahia2021low}. Generally speaking, emissions, matters of equity and accessibility are closely connected to those around climate change, and the centralization of resources remains a major problem~\cite{mattoo2012equity, morgan2014new}.

\subsection{What is the order of magnitude of \carbondioxide~emissions produced by training ML models?}~\label{sec:ml-emissions}

\begin{figure}[htp]
  \includegraphics[width=\linewidth]{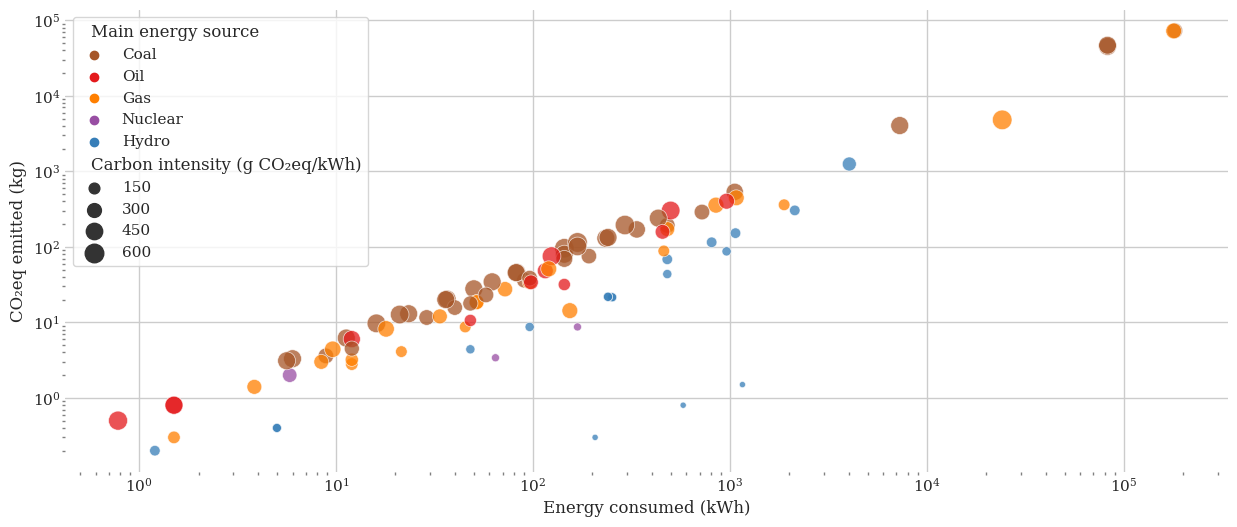}
  \caption{Estimated energy consumed (kWh) and \carbondioxide~(kg) by each model in the data set, plotted in a log-log scale. Colors indicate the principal energy source, and the size of the dot carbon intensity. While the relationship between energy and carbon emissions is mostly linear, the data show that models trained with less carbon-intensive energy (e.g. hydroelectric) emit orders of magnitude less carbon than those trained using more carbon-intensive energy (e.g. coal).}
  \label{fig:energysources}
\end{figure}

As explained in Section~\ref{methods}, there is a linear relationship between the energy consumed and the carbon emissions produced, with the energy source (discussed in the Section above) influencing the magnitude of this relationship. In Figure~\ref{fig:energysources}, we plot the energy consumed (X axis, logarithmic scale) and the \carbondioxide~emitted (Y axis, logarithmic scale) of every model in our data set, color-coded with the main energy source, which are the same as those presented in Table~\ref{table:energysource}. First, we can observe differences of several orders of magnitude in the energy used by models in our sample, ranging from just about 10 kWh to more than 10,000 kWh, which results in similar differences in the total quantity of \carbondioxide emitted. As expected, the relationship between energy consumed and carbon emitted is largely linear. However, Figure~\ref{fig:energysources} also shows that models trained with cleaner energy sources, such hydroelectricity, largely deviate from the main trend, with orders of magnitude less carbon emissions compared to models trained using coal and gas. In other words, models trained with low carbon-intensive energy sources, result in much less carbon emissions, \textit{ceteris paribus}.  

For instance, honing in on the central bottom portion of Figure~\ref{fig:energysources}, it can be seen that the models trained using hydroelectricity (the blue dots) are about two orders of magnitude lower in terms of carbon emissions than models that consumed similar amounts of energy from more carbon-intensive sources such as coal (in brown) and gas (in orange), given that the Y axis is on a logarithmic scale. Furthermore, the size of the dots varies as a function of the carbon intensity of the electricity grid used, illustrating two parallel groups of models, both exhibiting a largely linear trend, with the more carbon intensive models positioned higher than the lower carbon models for similar amounts of energy consumed. This further supports the analysis carried out in Section~\ref{sec:energymix}, suggesting that the primary energy source used for training ML models has a strong impact on the overall resulting emissions from model training, and that choosing a low-carbon energy grid can play a significant role towards reducing the carbon emissions of ML model training.

Besides the primary energy source, carbon emissions are a function of power consumed by the hardware used and the training time. The choice of hardware has a relatively small influence on the large variation of carbon emissions that we observe in our sample , given that the TDP ranges from 180 W to 300 W, while the carbon emissions span from $10^5$ kg\carboneq to even less than 10~kg\carboneq (see Section~\ref{appendix:hardware} of the appendix for further details). While using renewable energy can reduce up to 1,000 the carbon emissions for the same amount of energy used, the remaining factor responsible for the large variation in both energy and carbon emissions in our sample is therefore the training time.

\subsection{How do the \carbondioxide~emissions produced by training ML models evolve over time?}\label{sec:carbon-emissions}

Some recent analyses have predicted that the carbon emissions of our field will increase in the future, estimating that achieving further progress on benchmarks such as ImageNet will require emitting thousands of tons of~\carbondioxide~\cite{thompson2020computational}, whereas others have predicted a plateau in future emissions due to increased hardware efficiency and carbon offsetting~\cite{patterson2022carbon}. Therefore, one of the goals of our study was to observe the evolution of carbon emissions over time and study whether there are clear trends. Given that the papers from our study span from 2012 to the present time, we aimed to specifically compare whether new generations of ML models from our sample consistently used more energy and emitted more carbon than previous ones.

\begin{figure}[htp]
  \includegraphics[width=\linewidth]{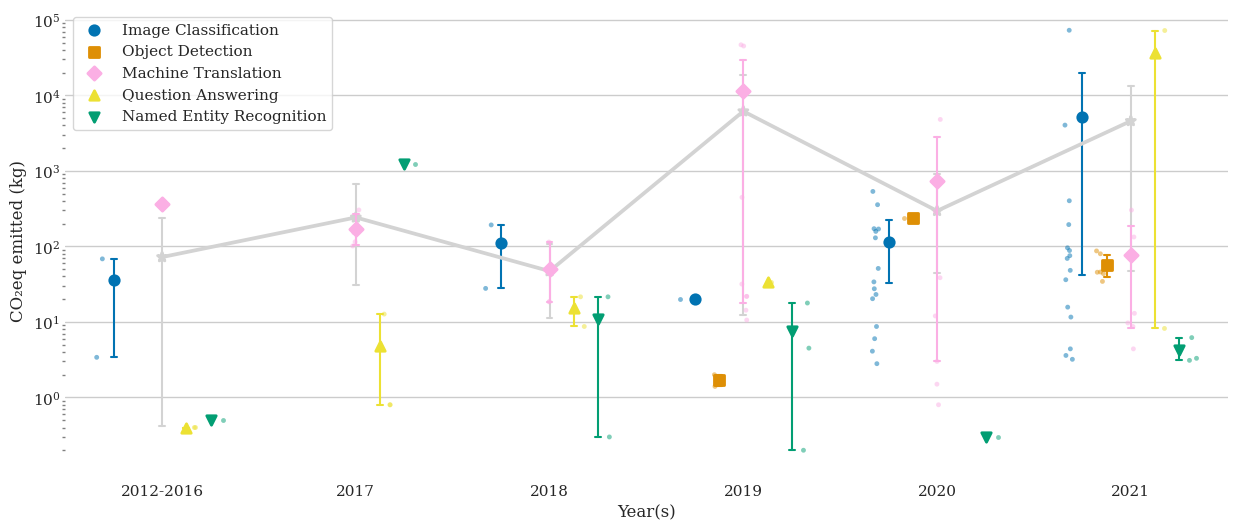}
  \caption{\carbondioxide~emitted (in kg) by the all models included in the data set, on a logarithmic scale. Each small marker corresponds to a model and the large markers indicate the 99 \% trimmed mean within each task and year(s) of publication. The error lines cover the bootstrapped 99 \% confidence intervals. The gray line corresponds to the average over all tasks.}
  \label{fig:co2emitted}
\end{figure}

In Figure~\ref{fig:co2emitted}, we show the carbon emissions emitted by every model from our sample, disaggregated by task and by year. While we cannot claim that the models and papers in our data set are fully representative of the whole machine learning field, a sample of 95 models spanning 9 years can offer interesting insights. The first observation, related to the conclusions from the sections above, is that there is a large variability in the carbon emissions from ML models. Second, we do not observe a consistent trend by which carbon emissions have systematically increased for each individual task. This is the case, for instance, of image classification models (in blue) and question answering models (in yellow) from our sample. However, the carbon emissions of machine translation models have peaked in 2019 and has since decreased. 

If we look at the aggregated data from all tasks (grand average curve, in light gray), we can observe that overall, the carbon emissions per model have increased by a factor of about 100 (two orders of magnitude) from 2012 to recent years, with slight fluctuations, as in 2020. It is important to note that the vertical axis of Figure~\ref{fig:co2emitted} is on a logarithmic scale, in order to reflect the non-linearity introduced by the much larger models from recent years, even though they do not represent a majority in the sample. In fact, the last three years of our sample (2019-2021), have seen models that have emitted orders of magnitude more carbon than before: e.g. there are several vertical outliers in tasks such as Image Classification (shown in blue) and Question Answering (in yellow) that have set new records in terms of the total amount of emissions produced by model training, responsible for about $10^{4}$ and $10^5$ kilograms of~\carboneq. There are several possible explanations for this, ranging from the widespread adoption of Transformers, which are using increasing amounts of both labeled and unlabeled data~\cite{vaswani2017attention}, as well as computationally-expensive techniques such as NAS~\cite{zoph2016neural}, which result in more carbon emissions~\cite{strubell2019energy}. It is hard to disentangle the influence of different factors on the overall carbon emissions of ML models, as well as the relative contributions of different parts of the pre-training and fine-tuning process -- this requires further work, which we discuss in Section~\ref{conclusion} -- however, it is worth noting the evolution of emissions in recent years, among the papers of our sample.

\subsection{Does more energy and \carbondioxide lead to better model performance?}~\label{sec:task-progress}

A final perspective from which we analyze the carbon emissions of ML models is by comparing the amount of carbon emitted by models to their performance on benchmark tasks such as image classification, machine translation and question answering.
We compare the emissions of the models from our sample and their performance on four tasks: image recognition on ImageNet~\cite{deng2009imagenet} (35 models), machine translation for English-French and English-German on the 2014 WMT Translation tasks~\cite{bojar2014findings} (30 models), question answering on the SQuAD 1.1 dataset~\cite{rajpurkar2016squad} (10 models), and named entity recognition on the CoNLL 2003 dataset~\cite{sang2003introduction} (11 models)~\footnote{We also had data from a fifth task, object detection, which is represented in Table~\ref{table:papers} and Figure~\ref{fig:co2emitted}, but we did not have enough distinct data points to enable a meaningful comparison.}. Our goal with this analysis is to validate whether, generally speaking, the more carbon-intensive models from our sample achieved better performance on common benchmarks compared to the models with less incurred emissions.

\begin{figure}[h!]
\centering
  \includegraphics[width=\linewidth]{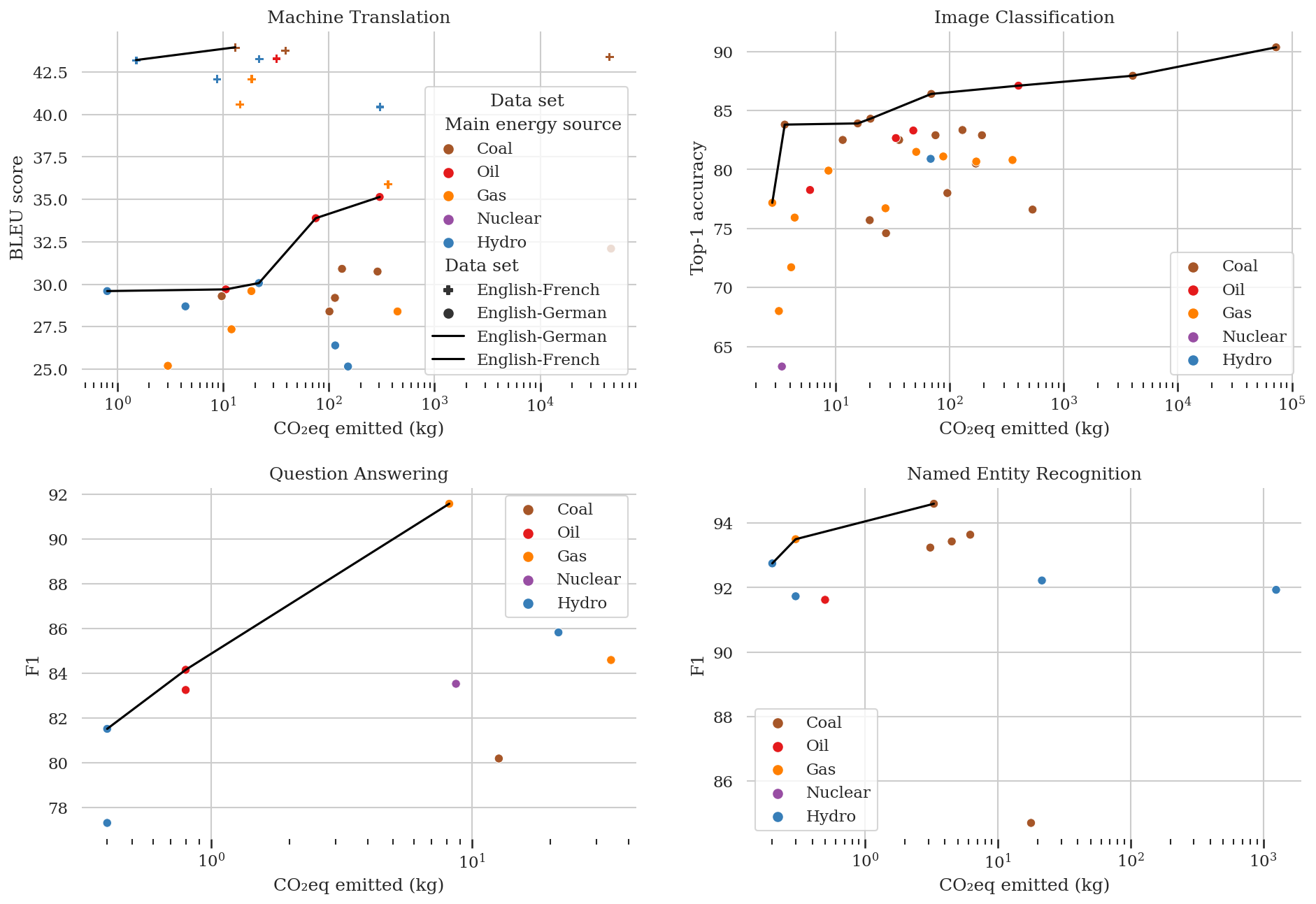}
  \caption{Comparison of the accuracy achieved by each model trained on \textbf{Machine Translation} (top left, evaluated using BLEU score on the English-French and English-German WMT datasets), \textbf{Image Classification} (top right, measured using Top-1 accuracy on ImageNet), \textbf{Question Answering} (bottom left, evaluated using F1 score on SQuAD v.1) and \textbf{Named Entity Recognition} (bottom right, evaluated using F1 score on the CoNLL dataset) and the \carbondioxide~emitted for training models. The black curves correspond to the Pareto fronts given the data, that is data points under the line are sub-optimal in terms of performance and \carbondioxide emitted.Note that the x axis is in logarithmic scale. }
  \label{fig:carbon_performance}
\end{figure}

Figure~\ref{fig:carbon_performance} shows the performance of the models in these four tasks and the associated carbon emissions; we also represent the theoretical Pareto front given the data, which corresponds to the set of Pareto-efficient solutions based on our data. We can think of the Pareto front of our metrics, the black line in the figures, as the curve connecting the models that achieved the best accuracy for a given amount of \carboneq emissions. In other words, all the data points under the Pareto lines correspond to models that obtained lower accuracy than other models in the sample despite producing the same or more carbon emissions. 

Based on the comparison between carbon emissions and performance, we can observe that the only task in which better performance accuracy has systematically yielded more \carbondioxide~is image classification on ImageNet, seen on the top right subplot of Figure~\ref{fig:carbon_performance}. Still, the relationship is far from being highly correlated (especially given that that the x-axis in on a logarithmic scale). For example, out of the 35 models analyzed, the top two models in terms of performance are also the most carbon-emitting. However, the third most carbon-intensive model is on the lower end of the performance (achieving ~76 \% accuracy), and we also see low-emitting models on the higher end of performance. 

For other tasks, the trend is even less clear -- for instance, for the 30 models evaluated on the WMT translation task (top left plot of Figure~\ref{fig:carbon_performance}), there is no clear link between \carbondioxide~emissions and BLEU score, for neither English-French or English-German -- although the WMT English-French task seems to incur more carbon emissions than the English-German one, which can be explained in part by the fact that the WMT English-French data set is almost 4 times larger than the English-German one, which can require a longer training time and thus a higher energy consumption. For the final two NLP tasks, question answering and named entity recognition, we have less data points (10 for the former and 11 for the latter), and the connection between carbon emissions and accuracy is very unclear. For both tasks, many models from both the high and low ends of the range of \carbondioxide~emissions achieve comparable performance on the SQuAD dataset (bottom-left plot) as well as the CoNLL dataset (bottom-right plot).

Despite the lack of clear correlation between carbon intensity and model performance, there are some interesting observations to be made based on Figure~\ref{fig:carbon_performance}. 
While we did not expect to see a strong link between these two factors, we find it worth noting that neither consuming more energy nor emitting more carbon seems to necessarily correlate with a higher accuracy, even in tasks such as Machine Translation, where Transformer models are largely seen to do better compared to other models~\footnote{We find a similar pattern between accuracy and energy consumption, which can be seen in Figure~\ref{fig:energy_performance} in the Supplementary Materials.}.

\section{Discussion and Future Work}~\label{discussion}

In the current section, we discuss the significance and the context of our analysis, its limitations, as well as promising directions for future work to improve the transparency of carbon emissions reporting in our field.

\subsection{Discussion of Results}

While the total carbon footprint of the field of ML is unclear due its distributed nature and the lack of systematic reporting of emissions in different settings, in the face of the climate crisis, it is important for the ML community to acquire a better understanding of its environmental footprint and how to reduce it~\cite{ligozat2021unraveling, patterson2021carbon}. Our study is the first analysis of the carbon emissions of a multitude of ML models from different perspectives ranging from energy source to performance. While our sample is only a small portion of the entire Machine Learning field, the carbon emissions associated to the models in our data set is significant: the total carbon emissions of the models analyzed in our study is about 253 tons of~\carboneq, which is equivalent to about 100 flights from London to San Francisco or from Nairobi to Beijing. While this may not seem like a large amount, the increase in emissions in recent years -- from an average of 487 tons of ~\carboneq~ for models from 2015-2016 to an average of 2020 tons for models trained in 2020-2022 -- as well as other trends that we observed in Section~\ref{sec:carbon-emissions}, indicate that the overall emissions due to ML model are rising. 

In Section~\ref{sec:results}, we have discussed that the main sources of variance in the amount of emissions associated to training machine learning models is due to the carbon intensity of the primary energy source and the training time, with the power consumption of the hardware having a smaller influence. In terms of training time, the models in our sample range from just about 15 minutes (total GPU/TPU time) up to more than 400,000 hours, with a median of 72 hours, pointing again to large variance in our sample. While the maximum of of 400,000 GPU hours (equivalent to about 170 days with 100 GPUs) in our sample seems very large, note that the total training time of GPT-3 was estimated to be over 3.5 million hours (14.8 days with 10,000 GPUs)~\cite{patterson2021carbon}. Obviously, such long training times result in large amounts of carbon emissions, even with lower carbon intensity energy sources. By way of illustration, the model with the longest training time in our sample would have reduced by about 30 times the carbon emissions had it used the grid with the lowest carbon intensity in our sample, but it would have still resulted in over 1 ton of \carboneq. Also, generally speaking, we can see that the models at the higher end of the emissions spectrum tend to be Transformer-based model with more layers (as well as using techniques such as Neural Architecture Search to find optimal combinations of parameters), whereas simpler and shallower models such as convolutional neural networks tend to be on the lower end of the emissions spectrum. Given that Transformer architectures are increasing in popularity -- especially in NLP but also for several Computer Vision tasks -- having a better idea of their energy consumption, carbon emissions, and the factors that influence them is also crucial part of analyzing the current and future state of our field.

An important observation from our analysis is that better performance is not generally achieved by using more energy. In other words, good performance can be achieved with limited carbon emissions because the progress in recent years has brought the possibility to train machine learning models efficiently. Image Classification is the task in our sample in which we observed the strongest correlation between performance and emissions. However, even in this task we also observed that small increments in carbon emissions lead to large increments in top-1 accuracy (see the left-hand-side of Figure~\ref{fig:carbon_performance}). This highlights the availability of efficient approaches and architectures. 
 
\subsection{Limitations} \label{limitations}

The analyses that we have carried out and the insights that they have provided us are useful towards a better understanding of the overall carbon emissions of ML model training. We are also aware of the limitations of our study: for one, we recognize that our sample is not fully representative of the field as a whole, given the diversity of models and architectures that exist and the speed at which our field is evolving. As we discussed in Section~\ref{methods}, despite our best efforts and several reminders, only 15\% of authors from our initial sample of 500 were willing to share relevant information with us. We also recognize that there are several factors that we are missing in order to be more precise in our estimation the carbon footprint of ML models: for instance, we do not have the necessary information regarding the Power Usage Effectiveness (PUE) of the data centers used for model training (i.e. the overhead used for heating, cooling, Internet etc.), as well as the real-time energy consumption of the hardware used for training. We also do not account for carbon offsets and power purchase agreements, which intend to bring computing centers closer to carbon neutrality and which are often taken into account by providers of cloud compute in their carbon accounting~\cite{google}. Despite this, the apples-to-apples carbon analysis that we carried out in the current study provides useful insights about the current state of carbon emissions in our field, as well as how this has evolved over time in the last 9 years.
 
Furthermore, while this study and much of the related work in this field has focused on estimating the carbon emissions of model training, there are many pieces of other overall carbon footprint of our field which are still missing: for instance, the carbon emissions of tasks such as data processing, data transfer, and data storage~\cite{ligozat2021unraveling}, as well as the carbon footprint of manufacturing and maintaining the hardware used for training ML models~\cite{gupta2021chasing}, We are also lacking information regarding the carbon impact of model development and inference -- given that a model that is trained a single time can be deployed on-demand for millions of queries, this can ultimately add up to more emissions than those produced by the initial model training~\cite{luccioni2022estimating}. These are all directions for future research, which we discuss in more detail below.
 
\subsection{Future Work} \label{conclusion}

There is much interesting and exciting work to be done that would help us better understand the carbon emissions and broader environmental implications of ML. This includes:

\paragraph{Additional empirical studies.} There is still a lot of uncertainty around, for instance, the relative contribution of added parameters of ML to their energy consumption and carbon footprint, as well as the proportion of energy used for pre-training versus fine-tuning ML models for different tasks and architectures. Furthering this research can benefit the field both from the perspective of sustainability and overall efficiency.

\paragraph{Widening the scope of ML life-cycle emissions.} The overwhelming majority of work in carbon accounting for ML models has been limited to model training. However, both the upstream emissions (i.e. those incurred by manufacturing and transporting the required computing equipment) as well as the downstream ones (i.e. the emissions of model deployment) warrant further exploration and better understanding. 

\paragraph{Increased standardization and transparency in carbon emissions reporting.} As stated in Section~\ref{limitations}, we put in significant efforts in contacting authors and gathering data to carry out our study, and were still lacking much of the necessary information that we would have liked to have. While certain conferences such as NeurIPS are starting to include compute information in submissions in submission checklists, there is still a lot of variability in carbon reporting, and figures can vary widely depending on what factors are included. Having a more standardized approach, such as ISO standards, to reporting the carbon emissions of ML can help better understand their evolution.

\paragraph{Considering the trade-off between sustainability and fairness.} The environmental impacts of ML also come with consequences in terms of fairness, given the interplay between fairness and sustainability, most recently discussed by~\citet{hessenthaler2022bridging}. This includes, for instance, the consideration of the environmental impacts of ML approaches when benchmarking models~\cite{zhou2020hulk}, but also, conversely, considering the impact on robustness and bias of model distillation techniques that improve model efficiency~\cite{hooker2020characterising,xu2022can}. Generally speaking, given that many advances in ML from last years can be attributed to training increasingly deep and computationally expensive models, especially in fields such as natural language processing, it is important to be cognizant of the broader societal impacts of these models, be it from the perspective of their energy consumption~\cite{dodge2022measuring, bender2021dangers}, the 
attribution of computing resources~\cite{ahmed2020democratization, ahia2021low} or the influence of corporate interests on research directions~\cite{abdalla2021grey, birhane2021values}.
\\

While discussions regarding the carbon footprint of our daily lives has started to become more common in many communities, alongside increased awareness of how our lifestyle choices (such as the way we travel and the food we eat) contribute to carbon emissions, we are lacking much of the necessary information necessary to regarding the impacts of the models we train. We hope that our work encourages better practices and more transparency in reporting the computational needs of the models and details of the energy used, and that our study will be a meaningful contribution towards a better understanding of our impact as ML researchers and practitioners. 
 
\newpage
\bibliographystyle{ACM-Reference-Format}
\bibliography{bibliography}

\newpage
\appendix

\section{Supplementary Materials}

\subsection{Emails sent to authors} \label{appendix:emails}

\texttt{\textbf{Subject:} Information Request: Computing Infrastructure Used in your Paper}

\texttt{Hello,}

\texttt{My name is XXXX and I am a researcher working on the environmental impact of Machine Learning. I am trying to gather data regarding the carbon footprint of recent state-of-the-art research papers. This will help the ML community get a better idea of how much CO2 we are emitting when training models.}

\texttt{In order to help me on my mission, I was hoping you could give me more information about your paper entitled YYYY.}

\texttt{More specifically, could you tell me:} \\
\texttt{- Where it was trained? If it was on a local computing cluster, could you tell me the location of the cluster? And if it was trained on the cloud, could you indicate the provider and server region (e.g. "Microsoft Azure, us-east1")?}\\
\texttt{- What hardware you used}\\
\texttt{- The total training time of your models?}\\

\texttt{Thank you very much for this information,}

\texttt{XXXX}

\subsection{Information regarding training hardware} \label{appendix:hardware}

\begin{table}[h!]
\caption{The top 5 GPUs/TPUs used, the number of models that used them for training, the range of quantities that were used, and their Thermal Design Power (TDP).}
\begin{center}
\begin{tabular}{lcll}
\toprule
\textbf{Model} & \textbf{Number of models} & \textbf{TDP} & \textbf{Quantity used} \\
\midrule
Tesla V100  & 30 & 300~W & 1-128  \\
TPU v3      & 9  & 450~W & 1-1024 \\
RTX 2080 Ti & 8  & 250~W & 4-16   \\
Tesla M40   & 5  & 250~W & 8      \\
GTX 1080    & 4  & 180~W & 1-8   \\
\bottomrule
\end{tabular}
\label{table:tdp}
\end{center}
\end{table}
\vspace{-10pt}

In Table~\ref{table:tdp}, we represent the 5 most popular GPU and TPU models used in the papers we analysed, accompanied by the number of papers that used them, the range of quantities used, and their TDP. The Tesla V100 was by far the most popular piece of hardware, representing almost a third of the papers, followed by the TPU v3. The TDP of the hardware used in our paper sample also varies significantly, from 180W for models such as the GTX 1080 to 450W for the TPU v3 model, meaning that TPUs, on average, consume more energy during usage.
Looking at the number of GPUs and TPUs used for ML training in the papers that we surveyed, we can see that there is a large range in the quantity of GPUs/TPUs used for model training, with some models leveraging up to 1024 TPU v3s for training, while others utilize a single GTX 1080 GPU for varying amounts of time, which makes the total energy consumption vary significantly. We analyze the connection between energy usage and performance on different ML tasks in \S~\ref{sec:task-progress}, in order to determine whether higher energy consumption helps achieve better performance in different ML tasks.


\subsection{Energy Consumption by Task}

In Figure~\ref{fig:energy_performance} below, we plot the same four tasks as in Figure~\ref{fig:co2emitted}, representing the energy consumed instead of the \carbondioxide~emitted. We find largely similar trends as the ones we describe in Section~\ref{sec:task-progress}, with better performance on tasks like machine translation and image classification not necessarily being contingent on higher energy consumption.

\begin{figure}[h!]
  \includegraphics[width=0.95\linewidth]{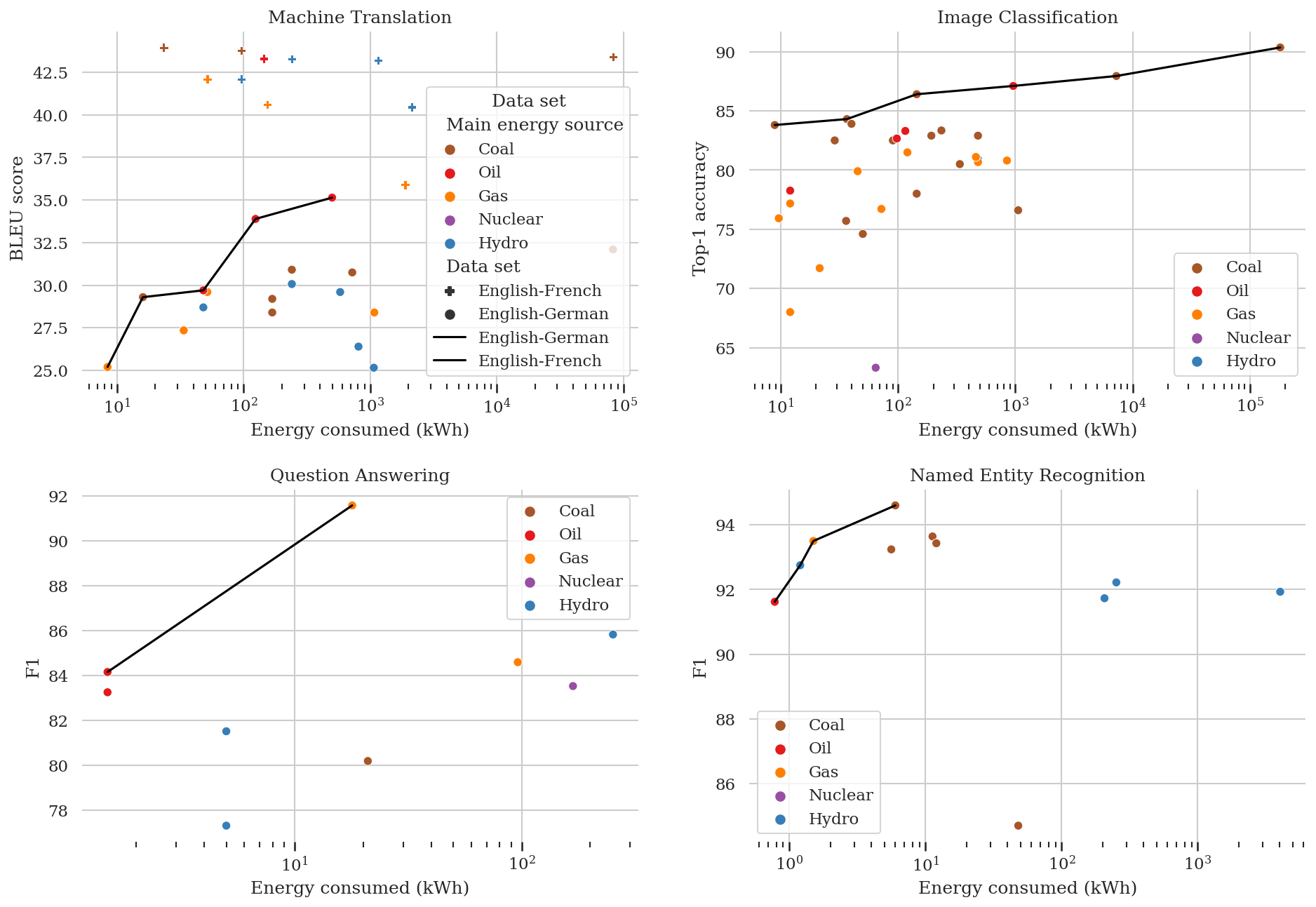}
  \caption{Comparison of the performance achieved by each model trained on Machine Translation tasks (BLEU score) and Image Classification (top-1 accuracy), and the energy consumed.}
  \label{fig:energy_performance}
\end{figure}
\clearpage
\subsection{Carbon intensity over time}

\begin{figure}[ht]
    \centering
    \begin{subfigure}{\linewidth}
      \includegraphics[width=\textwidth]{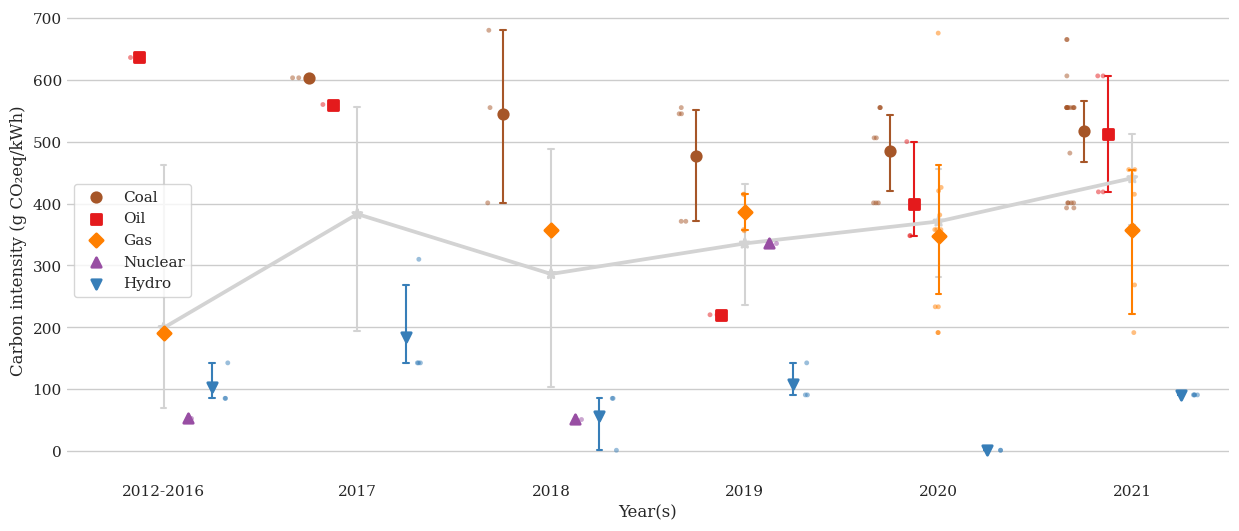}  
      \caption{Carbon intensity of the models per year and energy source.}
    \end{subfigure}
    \hfill
    \begin{subfigure}{\linewidth}
      \includegraphics[width=\textwidth]{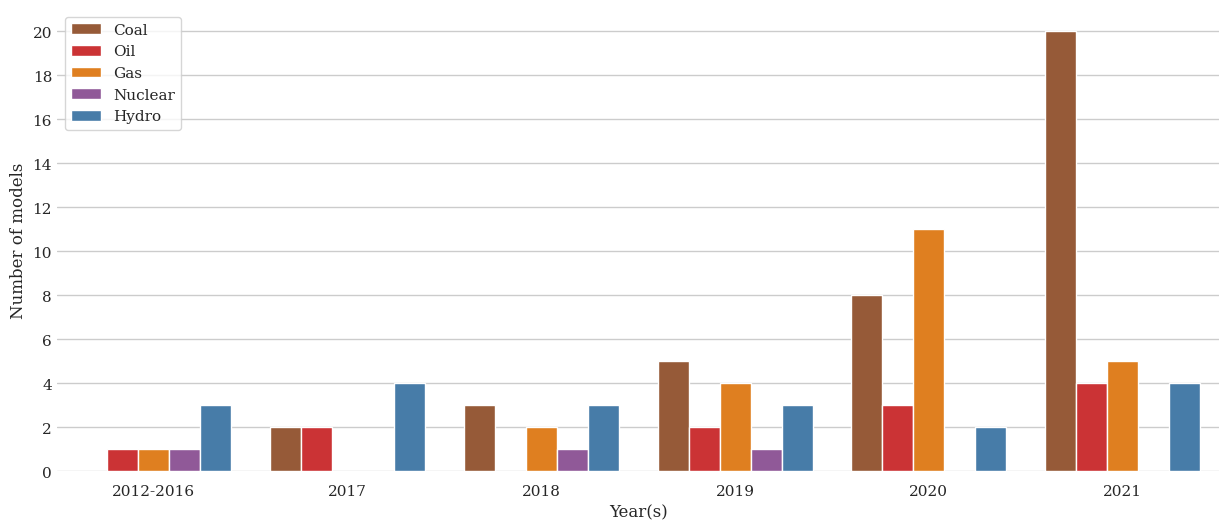}  
      \caption{Number of models trained with each energy source per year.}
    \end{subfigure}
\caption{Carbon intensity and energy sources over the years.}
\label{fig:carbon_intensity_time}
\end{figure}

In Figure~\ref{fig:carbon_intensity_time}, we plot the evolution over the years of the carbon intensity of the energy grid for each model, as well as the number of models trained with each energy source. We observe that, despite the need to address the climate crisis by using cleaner energy sources, there has not been a decrease in neither the average carbon intensity nor the number of models trained with cleaner energy. On the contrary, we do observe a stark increase of models trained with coal.

\end{document}